\documentclass{article}
\usepackage{ijcai24}
\usepackage{array}
\usepackage{times}
\usepackage{soul}
\usepackage{url}
\usepackage{tabularx}
\usepackage[table]{xcolor}
\usepackage[hidelinks]{hyperref}
\usepackage[utf8]{inputenc}
\usepackage[small]{caption}
\usepackage{graphicx}
\usepackage{amsmath}
\usepackage{amsthm}
\usepackage{booktabs}
\usepackage{algorithm}
\usepackage{algorithmic}
\usepackage[switch]{lineno}
\usepackage{amsfonts}       % blackboard math symbols
\usepackage{nicefrac}       % compact symbols for 1/2, etc.
\usepackage{microtype}      % microtypography
\usepackage{xcolor}         % colors
\usepackage{amsmath}         
\usepackage{makecell}
\usepackage{amssymb} % for \downarrow
\usepackage[utf8]{inputenc}
\usepackage{caption}  % for adjusting caption 
\usepackage{subfigure}
\usepackage{float}
\usepackage{subcaption}
\usepackage{stfloats}

\usepackage{graphicx}
\definecolor{best}{RGB}{255,0,0}
\definecolor{secondbest}{RGB}{0,0,255}
\usepackage{diagbox}
 \usepackage{amssymb}
\usepackage{algorithm}
\usepackage{algorithmic}
\usepackage{adjustbox}
\usepackage{siunitx} % For aligning numbers by decimal point
\usepackage{caption} % For improved table captions
\usepackage{booktabs} % For formal tables

 \usepackage{amssymb}
\usepackage{makecell}

\usepackage{listings}

 \usepackage{amssymb}
\usepackage{makecell}

\usepackage{listings}
\author{
Yangfan He$^{1,3}$
\and
Xinyan Wang$^2$
\and
Tianyu Shi$^2$
\affiliations
$^1$University of Minnesota - Twin Cities\\
$^2$University of Toronto\\
$^3$Henan RunTai Digital Technology Group Co., Ltd.\\
\emails
he000577@umn.edu,
 wangxinyan1123@gmail.com,
 tianyu.shi3@mail.mcgill.ca
}

\title{DDPM-MoCo: Advancing Industrial Surface Defect Generation and Detection with Generative and Contrastive Learning}

\begin{document}

\maketitle

%%%%%%%%% ABSTRACT
\begin{abstract}
The task of industrial detection based on deep learning often involves solving two problems: (1) obtaining sufficient and effective data samples, (2) and using efficient and convenient model training methods. In this paper, we introduce a novel defect-generation method, named DDPM-MoCo, to address these issues. Firstly, we utilize the Denoising Diffusion Probabilistic Model (DDPM) to generate high-quality defect data samples, overcoming the problem of insufficient sample data for model learning. Furthermore, we utilize the unsupervised learning Momentum Contrast model (MoCo) with an enhanced batchcontrastive loss function for training the model on unlabeled data, addressing the efficiency and consistency challenges in large-scale negative sample encoding during diffusion model training. The experimental results showcase an enhanced visual detection method for identifying defects on metal surfaces, covering the entire process, starting from generating unlabeled sample data for training the diffusion model, to utilizing the same labeled sample data  for downstream detection tasks. This study offers valuable practical insights and application potential for visual detection in the metal processing industry.
\end{abstract}

%%%%%%%%% BODY TEXT
\section{Introduction}
The defect detection of industrial products requires a large number of feature samples to train the network so that the model can distinguish individuals of the same type as the training samples. This presents two main challenges: acquiring a sufficient amount of sample data and annotating the distinguishing features in the data. In reality, obtaining a large dataset of product defects is difficult, and manually labeling each sample is tedious and monotonous . This study focuses on defect detection in industrial products, specifically on the surfaces of precision aluminum plates. This task presents a particular challenge due to the efforts of manufacturing enterprises to reduce the incidence of product defects to a minimum\cite{yang2020using}.
\begin{figure}[htbp]
    \centering
    \includegraphics[width=0.8 \linewidth]{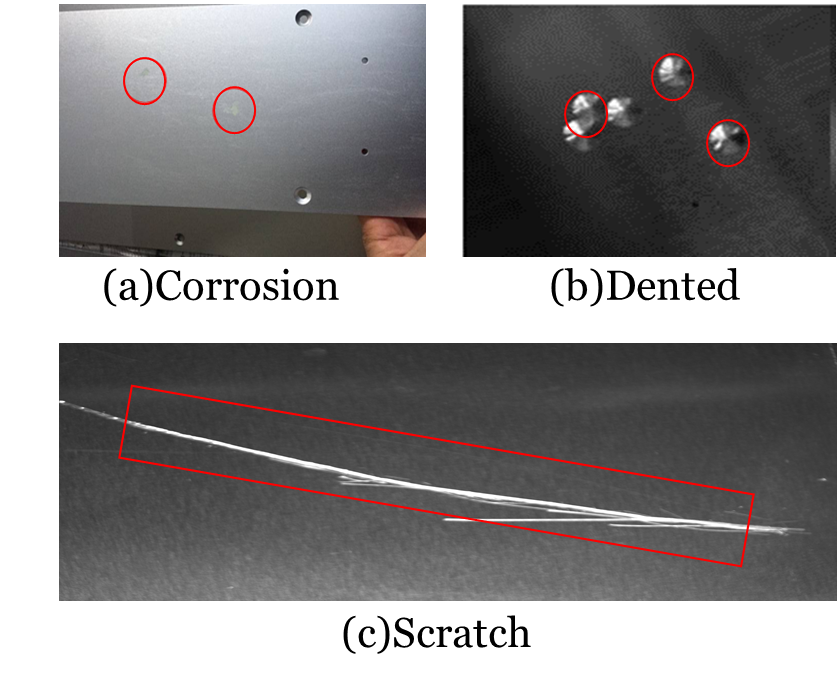}
    \caption{Examples of the three most common defect types on metal surfaces: (a) shows the aluminum plate with corrosion, (b) depicts a self-made aluminum plate with some dents, and (c) displays homemade scratched aluminum plate.}
    \label{fig:exampleclass}
\end{figure}
In general, metal surface defects include corrosion, cracks, dents, scratches, ink marks, and brittleness. These defects not only affect the appearance of the product but also indicate potential quality issues \cite{tao2018automatic}. In this research, we mainly focus on three surface defects that impact high-end aluminum products: dents, scratches, and corrosion. Dent is a single or multiple non-smooth depression defects produced by mechanical collision of aluminum plate, which is very destructive to the surface of the material. The scratch is a common defect in the processing, storage, and transportation of aluminum plates, caused by friction or scratching on the surface, and scratches damage the oxide film and the aluminum cladding layer, reducing the corrosion resistance of the material. The specific characteristics of corrosion are a little bit of white or black spots on the surface of the aluminum plate, which is caused by the production, packaging, transportation, storage process contact with acid or water. If the corrosion defects are not found and treated in time, they will not only make the surface of the aluminum plate lose its luster, but also reduce the corrosion resistance and comprehensive performance of the material \cite{vasagar2024non}.
as shown in Figure~\ref{fig:exampleclass}.

To be more specific, firstly, we utilize the probabilistic diffusion model (DDPM) to generate defect samples \cite{ho2020denoising}. Then, we combine the MoCo model with momentum contrast learning for efficient training without annotations \cite{he2020momentum}. Specifically, only a single sample is required for each type of defect in the diffusion model. The newly generated samples from the model are then mapped to the real data space. Once a large dataset is obtained, this paper utilizes the MoCo model for training without labeling data \cite{chen2020simple}. The model compares each individual sample with all other samples and continuously iterates and optimizes to capture inherent high-level structural characteristics of each sample data. Subsequently, it classifies samples of similar types by training a linear classifier. The proposed method DDPM-MoCo, potentially greatly mitigates the above issues,   Our main contribution includes:
\begin{itemize}
\item We developed the DDPM-MoCo model, combining Denoising Diffusion Probabilistic Models and Momentum Contrast for enhanced defect detection.
\item We addressed sample data scarcity by generating high-quality defect data and utilizing unlabeled samples for training.
\item We propose a novel batch contrastive representation loss that can enhance feature representation. 
\item Our experimental results demonstrate effectiveness in metal surface defect generation, offering a practical solution for industrial visual inspection tasks.
\end{itemize}

\section{\textbf{Related work}}
\noindent\textbf{2.1 Synthesis Defect Generation}\\
Due to a shortage of defective samples, many researchers ~\cite{schluter2022natural,zavrtanik2021draem,defard2021padim,yang2023memseg} have to make fake defects to enhance anomaly detection (AD) performance. They often do this by adding unusual pixels to normal images as a basic way to change the data. Baseline methods like CNN with cutout~\cite{devries2017improved} and the cutpaste approach~\cite{li2021cutpaste} randomly cut regions from normal images and paste them into "incorrect" places as artificial defects, creating artificial anomalies. Meanwhile, other techniques such as Crop\&Paste~\cite{lin2021few} and PRN~\cite{zhang2023prototypical} enhance anomaly detection (AD) by extracting real defect zones from flawed images and transferring them to flawless ones. While these methods outperform conventional one-class approaches, they lack the ability to generate novel defect patterns, potentially leading to overfitting issues. To increase the variations of anomalies, techniques like DeSTseg~\cite{zhang2023destseg}, MegSeg ~\cite{yang2023memseg}, DRAEM~\cite{zavrtanik2021draem} and ReSynthDetect~\cite{niu2023resynthdetect} include extra datasets combined with Berlin noise to create fabricate defects, but the distribution of these defects may not accurately reflect actual defects and thus there is considerable uncertainty regarding the performance gain. With the recent adavancement of AIGC(AI Generated Contents), some efforts have been made to create more realistic and diverse simulated defects. Generative Adversarial Network(GANs)~\cite{radford2015unsupervised,goodfellow2020generative,Niu2020defect} have been notable in research for their capability to generate high-fidelity industrial defect images, but they often present challenges such as difficult training processes and potential mode collapse. Moreover, 
Auto-regressive models~\cite{bu2010detection,kulkarni2019automated} used in industrial image generation excel at detailing defect detection by sequentially processing pixels or blocks, nonetheless, they notably struggle with computational intensity and long-range dependencies, impacting their efficiency and scalability. Additionally, Variational Autoencoders (VAEs)~\cite{rescsanski2023anomaly,lu2023defvae} are favored for their robustness in handling different data distributions, yet they often produce blurry outputs and struggle in generating high-quality industrial defect images. Furthermore, Normalizing Flows~\cite{rudolph2021same,rudolph2024industrial,kobyzev2020normalizing} are used in industrial defect image generation, offering precise modeling through exact likelihood calculations, however, they face challenges with high-dimensional data and require complex architectures that increase computational demands, which may limit their effectiveness compared to other models like GANs or VAEs. Conversely, Denoising Diffusion Probabilistic Models (DDPM)~\cite{ho2020denoising,lin2024machine,honghui2023metal} have demonstrated significant promise in accurately and diversely generating industrial defect images, which is critical for advancing anomaly detection in industrial settings, thereby contributing to the development of more sophisticated diagnostic tools.\\
\noindent\textbf{2.2 Improving Industrial Defect Detection via Contrastive Learning as Pretask Tasks}\\
Self-supervised learning methods typically emphasize pretext tasks. The term "pretext" implies that the task serves primarily to learn data representations rather than as the ultimate objective. Various pretext tasks often utilize contrastive loss functions, with instance discrimination~\cite{wu2018unsupervised} methods associated with exemplar-based tasks~\cite{dosovitskiy2014discriminative} and NCE~\cite{gutmann2010noise}. In defect detection, self-supervised learning methods are crucial due to the time-consuming nature and potential oversight of minor defects in manual inspection. They utilize self-supervised learning as pre-training tasks, leveraging contrastive loss functions crucial for handling industrial defect images lacking explicit labels \cite{xie2021propagate,chen2020big,oord2018representation}. This approach enables accurate evaluation of sample similarity in the data representation space, effectively distinguishing between normal and abnormal conditions \cite{he2020momentum,hjelm2018learning,wu2018unsupervised}. Specifically, in industrial defect detection, contrastive loss functions assist the algorithm in capturing subtle disparities in industrial images, such as variations in color, texture, or shape, thereby enhancing the discrimination of potential defects through improved latent representations.
\section{\textbf{\textbf{Methodologies} } }
\begin{figure*}
    \centering
    \includegraphics[width=0.8\linewidth]{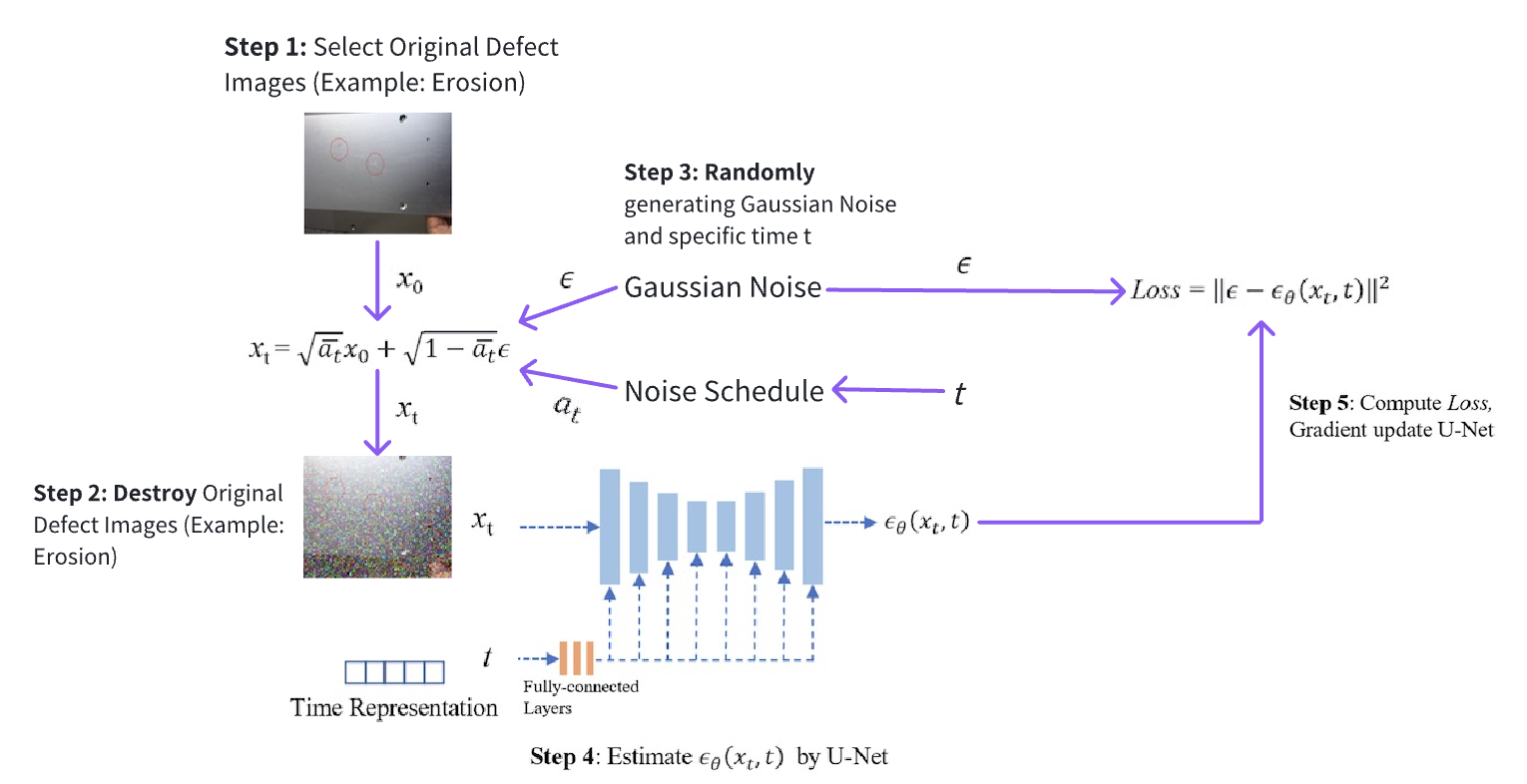}
    \caption{\textbf{DDPM training process.} We train diffusion models to generate new images by gradually adding Gaussian noise to simulate damage and create noise maps. Specifically, we input images with noise at time $t$ in UNet to predict the noise generated at time $t-1$, and update the UNet parameters based on the differences in Gaussian noise between time $t-1$ and $t$.}
    \label{fig:fig44}
\end{figure*}
Our work is based on~\cite{chen2020improved,ho2020denoising}. The model we are using is U-Net architecture from guided diffusion, and the attention mechanism is also applied in DDPM. In this section, we illustrate step by step how we apply a diffusion process to metallic surface defect generation, and we demonstrate the application of a MoCo process for model training performed with unlabeled sample data.

\subsection{Evolving Image Generation: From Deep Models to DDPM}
 Denoising Diffusion Probabilistic Models (DDPM)~\cite{ho2020denoising}, a particularly influential class of deep generative models, generate high-fidelity images from random noise, rivaling or surpassing traditional models like GANs and VAEs. The recent surge in advanced diffusion models is largely due to DDPM's foundational principles, which include a forward process, a backward pass, training methodologies, and practical applications, providing a comprehensive framework for image generation that excels in versatility and performance.

\subsubsection{Forward process (diffusion process)}The original defect image \textit{x}\textsubscript{0} is transformed into \textit{x}\textsubscript{T} by gradually adding Gaussian noise, so as to achieve the purpose of distorting the image. The forward process can be formulated as follows: 
\begin{equation}
x_t = \sqrt{a_t}x_{t-1} + \sqrt{1-a_t}\epsilon_{t-1}
\label{eq : forward_iter}
\end{equation}
where \(\{a_t\}_{t=1}^{T}\) is a pre-defined hyper parameter, called Noise schedule, which often includes columns with very small values. \(\epsilon_{t-1} \sim N(0,1)\) is Gaussian noise. 

As pointed out in \cite{ho2020denoising}, this iterative noising process can be simplified through an identity transformation, resulting Equation \ref{eq : forward_process}: 
\begin{equation}
x_t = \sqrt{\bar{a_t}}x_0 + \sqrt{1-\bar{a_t}}\epsilon
\label{eq : forward_process}
\end{equation}Here \(\bar{a_t}\) is also a super parameter set within the noise schedule, \(\epsilon \sim N(0,1)\) is an equivalent Gaussian noise. So the forward process can be depicted by Equation~\ref{eq : forward_iter} or~\ref{eq : forward_process}, and the Equation~\ref{eq : forward_iter} destroys an input image step by step, but ~\ref{eq : forward_process} do it in one step.

% \begin{equation}
% x_t = \sqrt{\bar{a_t}}x_0 + \sqrt{1-\bar{a_t}}\epsilon
% \label{eq : forward_process}
% \end{equation}
\subsubsection{Backward process (denoising process) }The reverse process is to gradually restore the damaged \textit{x}\textsubscript{T} to \textit{x}\textsubscript{0} by estimating the noise and iterating many times. 
The backward process can be formulated as follows:
% 这个地方应该是a_t bar，前面a_t bar=a_1*a_2* ...* a_t的相乘可能需要3.2.1中定义一下
\begin{equation}
x_{t-1} = \frac{1}{\sqrt{\bar{a_t}}}x_t - \frac{\sqrt{1-{\bar{a_t}}}}{\sqrt{\bar{a_t}}}\epsilon_\theta(x_t,t) + \sigma_t
\label{eq : backward_process}
\end{equation}
The Figure~\ref{fig:fig44} illustrates the training process of DDPM.
Since the real noise \(\epsilon\) in Equation~\ref{eq : forward_process} can not be used in the restoration process, the key to DDPM is to train a model \(\epsilon_\theta(x_t,t)\) that estimates the noise from the actual \(x_t\) at time \(t\). Here \(\theta\) is the training parameter of the model. \(\sigma_t \sim \textit{N}(0,1)\) is  Gaussian noise indicating the difference between the estimate and the actual. In DDPM, U-Net serves as a framework for estimating noise.

\subsubsection{Model training }From the above, we know the key to DDPM is to train a model \(\epsilon_\theta\left(x_t,t\right)\)and it should be made to predict \(\hat{\epsilon}\) close to the \(\epsilon\) that is actually used for destruction. So L2 distance is a good way to describe the similarity and the Loss is formulated as:
\begin{equation}
Loss =\left \|\epsilon -\epsilon_{\theta}(x_{t},t)\right \|^{2} 
     =\left \|\epsilon -\epsilon_{\theta}(z_{t},t)\right \|^{2}
     \label{eq : loss}
\end{equation}

\subsection{Enhancing Data Representation and Extraction with MoCo}

\subsubsection{Momentum Contrast (MoCo) }
The MoCo algorithm~\cite{he2020momentum}  enhances unsupervised learning through a dynamic dictionary for contrastive learning and optimizes data representation using an adjusted Noise-Contrastive Estimation (NCE) loss~\cite{he2020momentum} , described as the original loss in subsequent sections of our paper (see Equation~\ref{eq : Nce_Loss}). It utilizes a temperature parameter \(\tau\) to differentiate a query \textit{q} from \textit{k\textsubscript{+}} among \textit{k}+1 options by comparing \textit{q} to one positive key \textit{k\textsubscript{+}} and \textit{K} negative keys. This method promotes learning from both matched and unmatched image views:
\begin{equation}
\mathcal{L}_q=-\log{\frac{\exp{(q \cdot k_+/\tau)}}{\sum_{i=0}^{K}\exp{(q \cdot k_i/\tau)}}}
\label{eq : Nce_Loss}
\end{equation} And MoCo enhances model flexibility and scalability by managing a dynamic queue that enqueues new mini-batches and dequeues the oldest, thereby decoupling dictionary size from mini-batch size and continuously updating the dictionary with unique mini-batch elements. The algorithm stabilizes the learning process through a momentum-based update mechanism, where the key parameters \(\theta_k\) are incrementally updated in relation to the query encoder parameters \(\theta_q\), which are updated via gradient descent.
\begin{equation}
\theta_k \gets m\theta_k + (1-m)\theta_q, \quad m \in [0,1)
\label{eq : mo}
\end{equation}
This update mechanism ensures gradual changes, minimizing feature discrepancies and thereby enhancing the stability and performance of the learning model.

\subsubsection{Enhanced Batch Contrastive Representation Loss}

Our DDPM model has a structure broadly similar to the aforementioned DDPM model, as formalized in Equation~\ref{eq : Dataset_Discrimination}.

\begin{equation}
\mathcal{L}_q=-\frac{1}{n}{\sum_{j=1}^{n}}\log{\frac{\exp{(q \cdot k_j/\tau)}}{\sum_{i=0}^{K}\exp{(q \cdot k_i/\tau)}}}
\label{eq : Dataset_Discrimination}
\end{equation}

Compared with the structure of Equation~\ref{eq : Nce_Loss}, our dataset-level discrimination (Equation~\ref{eq : Dataset_Discrimination}) referred to as the improved loss in subsequent sections, with \(n\) as the mini-batch size is improved by not performing element-wise multiplication and summing between each query \(q\) and its corresponding positive key \(k+\) (i.e., \(nc, nc \rightarrow n\) process), but by considering all positive key samples globally. Specifically, each query matrix performs standard matrix multiplication with all \(k+\) matrices (\(nc \times cn \rightarrow n \times n\)), generating a \(n \times n \times n\) three-dimensional matrix. We then average across the third and second dimensions, ultimately reducing it to \(n \times 1\). Additionally, our loss function calculates the global average of the product of each query with all positive samples, and also averages each individual product result. This approach not only reflects the characteristics of global positive keys but also enhances the robustness and expressiveness of the model through averaging. The treatment of negative samples remains unchanged.
% \begin{figure*}
%     \centering
%     \includegraphics[width=0.75\linewidth]{image_13.png}
%     \caption{Basic structure of the MoCo model highlighting the FIFO (First In, First Out) queue mechanism.}
%     \label{fig:fig5}
% \end{figure*}

% \begin{figure*}
%     \centering
%     \includegraphics[width=0.75\linewidth]{image_11.png}
%     \caption{Architecture of the DDPM-MoCo model, integrating DDPM Training (see Figure 3) with MoCo Training (refer to Figure 4).}
%     \label{fig:fig6} % Updated label to fig:fig6 for uniqueness
% \end{figure*}

\section{Experiments}
\subsection{Experiment Setting}
1) \textit{\textbf{Datasets}}\\
Our experimental setup utilized a GS3-U3-50S5M-C camera to photograph the surface of an aluminum plate, segmented into six $80mm\times 80mm$ blocks. We introduced three types of defects, dent, scratches and errosion using an electric drill on each block, alongside identifying products with common production defects. 
\begin{figure}
    \centering
    \includegraphics[width=0.9 \linewidth]{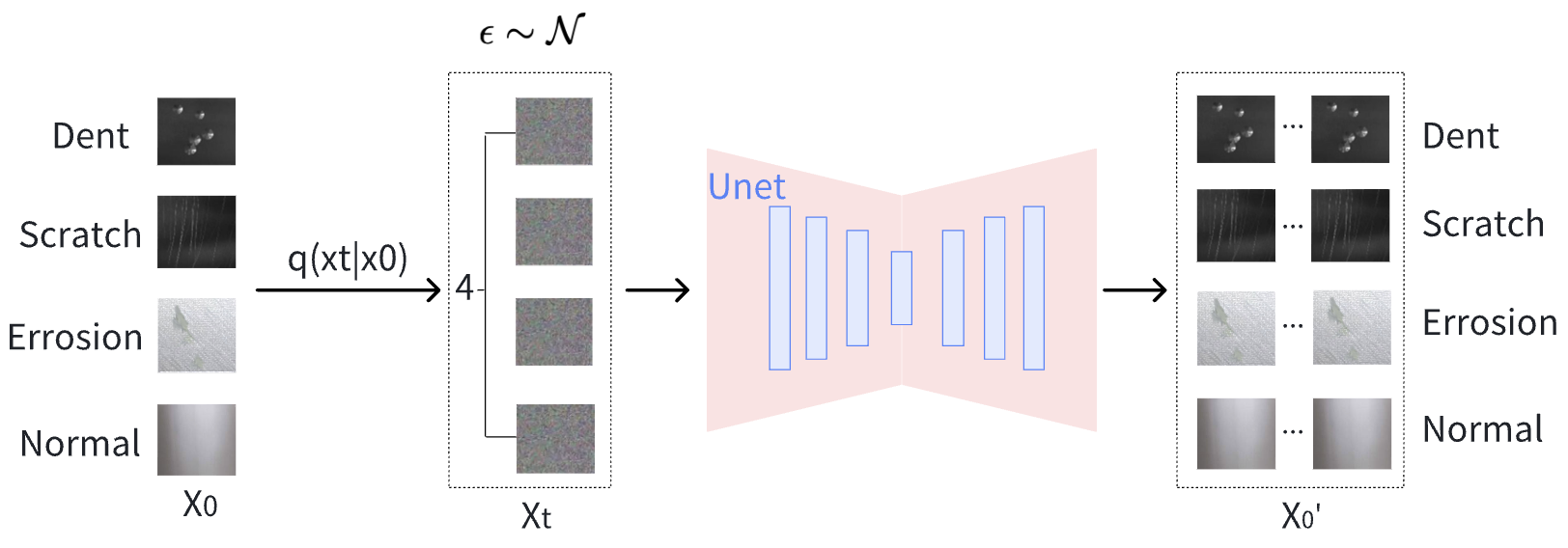}
    \caption{\textbf{Data augmentation.} We used four diffusion models to learn noise patterns across four image categories (dent, corrosion, scratch, smooth), utilizing limited samples of each defect type obtained from initially augmented images through horizontal flipping and edge padding for few-shot learning. Adjusting seed values, we generated a large quantity of random noise images and fed them into the four trained models to produce additional defect images across various categories.}
    \label{fig:fig5}
\end{figure} 
We have four types of images in our collection: one type is of normal conditions, and the other three types feature metal defects—dents, corrosion, and scratches, with 50 images for each category. To expand our dataset, we first applied traditional data augmentation techniques such as random flipping, edge padding, and contrast enhancement to 200 collected sample images of four types. This increased the sample size to 1600 images (400 images for each category). Subsequently, we deployed four untrained diffusion models, each embedded with \( t = 1000 \) time embeddings, designed specifically to train on four types of enhanced images including. After training, each model processed a substantial volume of random noise images, producing a significant collection of images across the four categories respectively. Regarding dataset partitioning, we allocated 80\% to the training set and 20\% to the test set, ensuring ample data for effective model training and performance evaluation. Consequently, 80\% of the expanded dataset of 20,000 images (5000 images for each category) were distributed across numerous batches, each with a size of 32, for subsequent contrastive learning training.\\
2) \textit{\textbf{Utilizing Backbones:}}\\
 \textbf{ResNet50} with 'skip connections' is used for both $encoder_q$ and $encoder_k$ in contrastive learning on four types of aluminum defects; after removing $encoder_q$, $encoder_k$ is modified with an average pooling and a fully connected layer, with its fixed backbone parameters refined via supervised learning on labeled images for classification.\\
 \textbf{ViT-B/16}~\cite{dosovitskiy2020image} uses a Transformer encoder to process images into patches for contrastive learning on aluminum defects, training $encoder_k$ with gradient descent and updating $encoder_q$ via momentum; after discarding $encoder_q$, $encoder_k$ gains a "classification token" and its neck and MLP head are precisely fine-tuned using labeled images in supervised learning, keeping backbone parameters fixed.\\
3) \textit{\textbf{Evaluation metric for anomaly detection:}}
\begin{enumerate}
    \item \textbf{Precision} : evaluates the classification accuracy of the four types in generated images, defined as the ratio of true positives to all predicted types. High precision indicates fewer false positives.
    \item \textbf{Recall}: assesses the model's ability to identify all four types of images, calculated as the ratio of true positives to total actual types, crucial for comprehensive type detection despite potential false positives.
    \item \textbf{AUC-PR(AP)}: The Area Under the PR Curve (AUC-PR) is a key metric for evaluating classification models. It summarizes the model's performance with a single scalar value by measuring the area under the PR Curve. A higher AUC-PR indicates better accuracy in defect detection, essential for industrial reliability.
\end{enumerate}
4) \textit{\textbf{Evaluation metric for Generation Quality:}}
\begin{enumerate}
    \item \textbf{Fréchet Inception Distance (FID)}: The FID quantifies the similarity between the distributions of generated and real images, serving as a crucial measure of the realism and quality of synthetic images. It assesses how closely generated images replicate the content and style of authentic images.
    \item \textbf{Inception Score (IS)}: The IS measures the clarity and diversity of generated images by analyzing model-predicted class probabilities. This metric evaluates the distinctiveness and variety of the images, ensuring that the generation process produces not only clear but also diverse outputs.
\end{enumerate}
% \begin{figure}
%     \centering
%     \includegraphics[width=0.6\linewidth]{image-3-2_11.png}
%     \caption{Laboratory workbench setup for data collection.}
%     \label{fig:fig6}
% \end{figure}

% This approach not only enhances the model's ability to generalize across various defect types but also aligns with industrial detection needs by preparing the model with a dataset reflective of practical scenarios.
\subsection{Implementation of DDPM-MoCo for defect generation}
\subsubsection{Implementation of DDPM}
The images from our dataset are cropped to small patches ($640\times 640$) and then resized to a smaller resolution ($512\times 512$) as input to the diffusion model. The processed data \textit{x}\textsubscript{t} and \textit{time} will be input into the U-Net main network of the diffusion model. The module Attn utilizes a linear attention mechanism~\cite{vaswani2017attention} to interact and reorganize input data, extracting key information while maintaining the original data dimension. This aims to generate more representative output The output of the model under the current batch input,  \(\epsilon_\theta\) in Equation~\ref{eq : loss}, can be obtained after multiple feature extraction and time information fusion.

%\begin{figure*}[htbp]
%    \centering
%    \includegraphics[width=1.0\linewidth]{image_04.png}
%    \caption{Overview of Denoising Diffusion Probabilistic %Model. (a) Training processes with the output predicting the %noise corresponding to timesteps. (b) Autoregressive sampling %processes based on equation 3 and forward processes Noise %schedule, with the begining of $x_T$ sampling from standard %Gaussian.}
%    \label{fig:fig13}
%\end{figure*}

\subsubsection{Improved training loss schedule}  The problem of infinite loss in the training process is addressed by proposing a dynamic model learning rate adjustment plan based on the training step, and we propose a learning rate adjustment scheme based on the cosine transformation, shown as Equation~\ref{eq : lr}, where \textit{current\_steps} is the number of epochs currently trained, and \textit{total\_steps} is the total number of epochs (epochs) of training. Obviously, due to the sine function, the learning rate of model training will gradually decreases from 1 to 0 during training, which ensures the stability of training to the greatest extent. 
\begin{equation}
    lr=\frac{1}{2}\left(1+cos{\left(\pi\frac{current\_steps}{total\_steps}\right)}\right)
    \label{eq : lr}
\end{equation}
\subsubsection{\textbf{Implementation of Momentum Comparison Model (MoCo)}}

The model uses three critical hyperparameters: dictionary size \(K\), momentum \(m\), and temperature \(\tau\) for the contrastive loss. The encoder \textit{f\textsubscript{q}} and encoder \textit{f\textsubscript{k}} handle matching samples and dictionary entries, respectively, using two distinct sets of backbone architectures. The first set employs the ResNet50 backbone architecture for both \textit{encoder\_q} and \textit{encoder\_k}, while the second set uses the Vitb-16 backbone architecture for these encoders. Our training set contains 16,000 samples. The training parameters were set as follows: \(K=16384\), \(m=0.999\), \(\tau=0.07\), a batch size of 32, and an initial learning rate of 0.03, which was dynamically adjusted using a cosine transformation (Equation \ref{eq : lr}). The training was conducted over 200 epochs using an NVIDIA GeForce RTX 3090 GPU.
\begin{figure*}[t]
    \centering
    \includegraphics[width=0.6\linewidth]{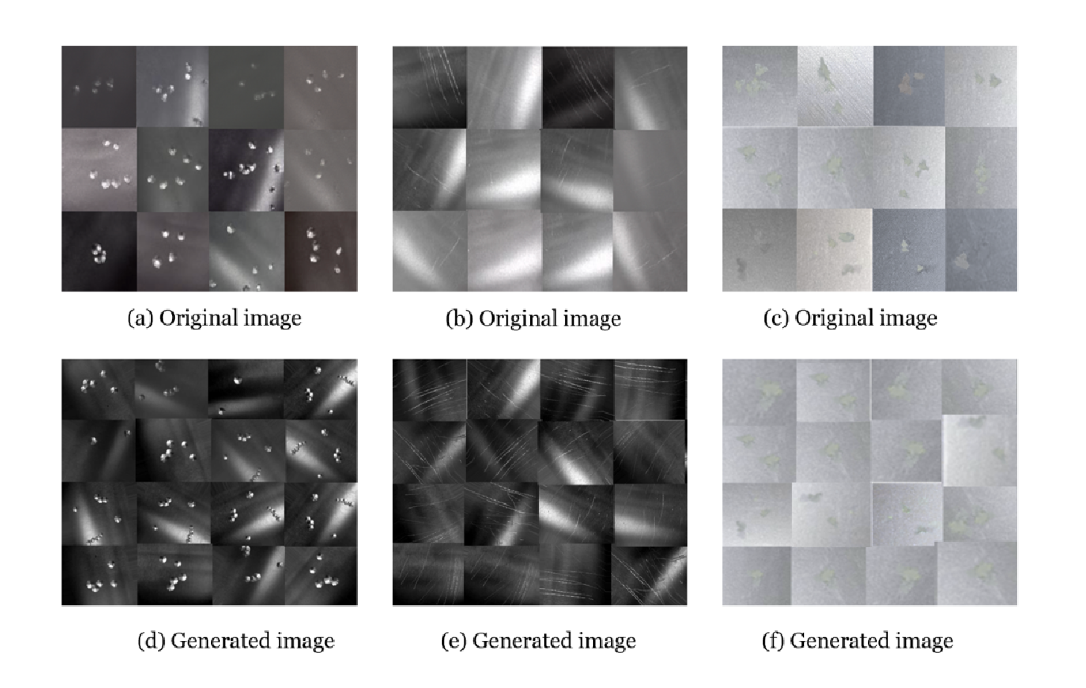}
    \caption{Original images and images generated by the diffusion model of the (d) dented defect, (e) scratch defect, and (f) corrosion defect. }
    \label{fig:fig14}
\end{figure*}
\subsection{Quantitative results}
\begin{table}[htbp]
    \centering
    \caption{FID and Inception Scores for the Augmented Dataset}
    \label{tab : data_generation}
    \begin{adjustbox}{width=0.35\textwidth, center}
    \begin{tabular}{@{}lccc@{}} % 使用c列类型使得内容居中
        \toprule
        {Case} & {Corrision} & {Dented} & {Scratch} \\
        \midrule
        FID Score & 2.17 & 2.85 & 1.94 \\
        Inception Score & 8.79 & 9.21 & 9.03 \\
        \bottomrule
    \end{tabular}
    \end{adjustbox} 
\end{table}

\begin{figure}[hbtp]
    \centering
    \includegraphics[width=0.6\linewidth]{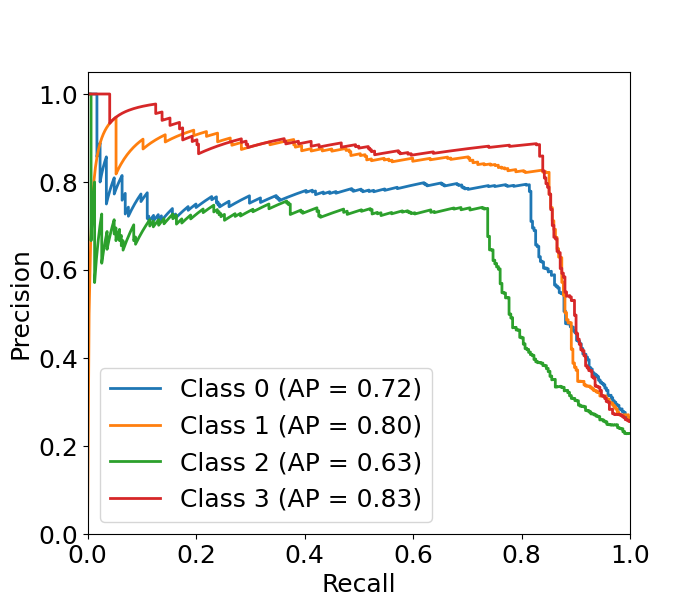}
    \caption{PR curve overview under orginal loss for Vitb-16}
    \label{fig:fig1}
\end{figure}
\begin{figure}[hbtp]
    \centering
    \includegraphics[width=0.6\linewidth]{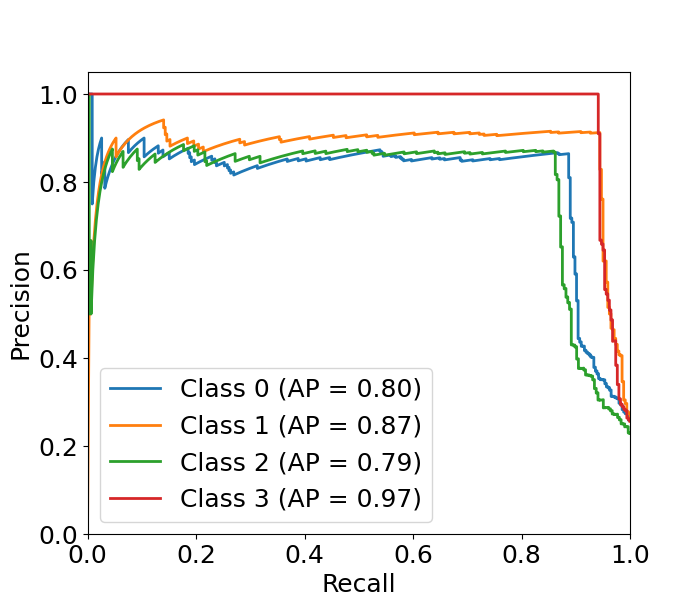}
    \caption{PR curve overview under improved loss for Vitb-16}
    \label{fig:fig2}
\end{figure}
% \begin{figure*}[ht]
%     \centering
%     \includegraphics[width=0.75\linewidth]{image_17.png}
%     \caption{Overview of the capabilities of four classic deep generative models in generating images with defects.}
%     \label{fig:fig1}
% \end{figure*}
\begin{figure}[hbtp]
    \centering
    \includegraphics[width=0.6\linewidth]{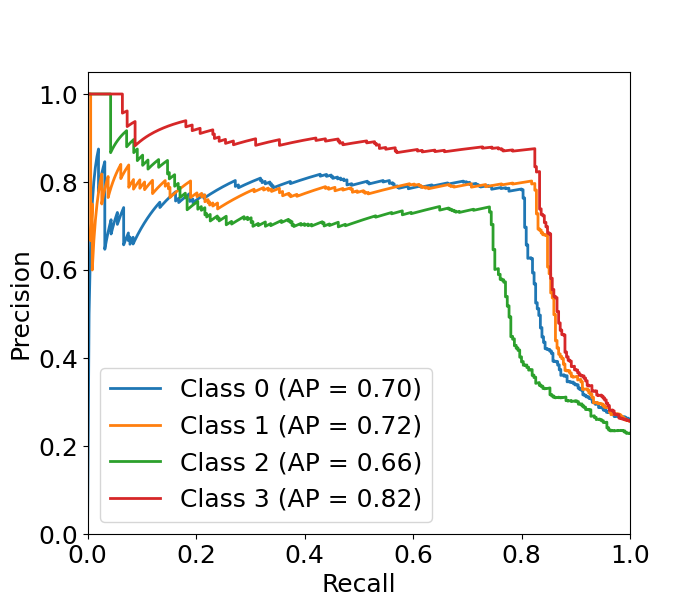}
    \caption{PR curve overview under orginal loss for Resnet-50}
    \label{fig:fig3}
\end{figure}
\begin{figure}[hbtp]
    \centering
    \includegraphics[width=0.6\linewidth]{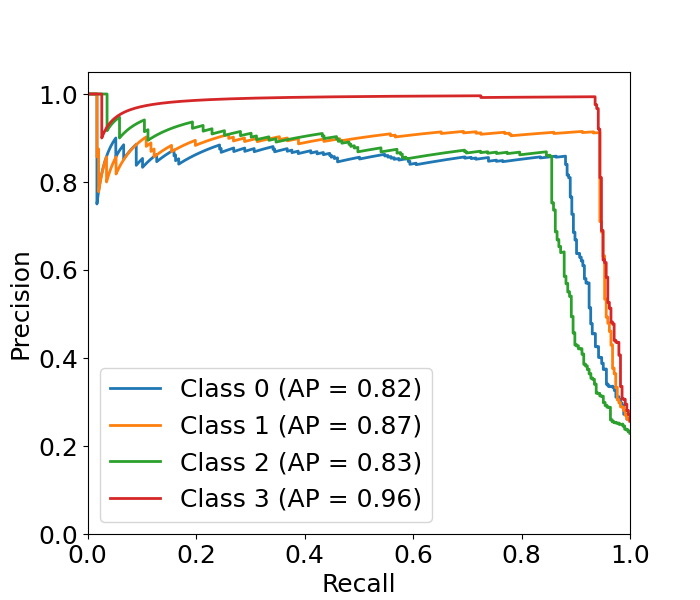}
    \caption{PR curve overview under improved loss for Resnet-50}
    \label{fig:fig4}
\end{figure}
\noindent\textbf{\textit{1) Image Quality:}}\\
Table ~\ref{tab : data_generation} showcases the FID and IS scores for both the original and extensively augmented datasets, emphasizing the maintained image quality. The substantial expansion of the augmented dataset is critical, particularly in scenarios where defect images are scarce, as it significantly enhances contrast learning. The low FID and high IS scores for the augmented data demonstrate that, despite a significant increase in the number of synthetic images, the quality and diversity of these images remain comparable to the original real images. This confirms that our data augmentation techniques effectively preserve image quality without compromise, even with a substantial increase in image quantity.\\
\noindent\textbf{\textit{2) Results under Original Loss and Improved Loss for Resnet-50 and ViT-B/16.}}\\
The dataset-level discrimination loss, detailed in Equation~\ref{eq : Dataset_Discrimination}, significantly boosts the performance of ViT-B/16 and Resnet-50 models, as shown by their PR curves. Initially, ViT-B/16 with the original loss (Equation~\ref{eq : Nce_Loss}) records AP values for classes 0 to 3—corrosion, dent, scratch, and smooth—as 0.72, 0.80, 0.63, and 0.83 respectively in Figure~\ref{fig:fig1}. Implementing our improved loss function raised these to 0.80, 0.87, 0.79, and 0.97 in Figure~\ref{fig:fig2}. Similarly, Resnet-50's initial AP values using the original loss were 0.70, 0.72, 0.66, and 0.82 (Figure~\ref{fig:fig3}), which improved to 0.82, 0.87, 0.83, and 0.96 with the improved loss (Figure~\ref{fig:fig4}). Particularly, the improved loss function most significantly enhanced Class 3's performance, boosting it from 0.83 to 0.97 for ViT-B/16 and to 0.96 for Resnet-50, probably due to its  simple attributes, which facilitate easier learning. The improved loss function aligns PR curves closer to the upper-right corner, indicating enhanced detection accuracy and robustness across classes, and highlights the models' dynamic performance sensitivity to threshold adjustments.

\begin{table*}[htbp]
\centering
\caption{AP Comparison for Dent (Dent-AP), Scratch (Scratch-AP), Erosion (Erosion-AP), and Smooth (Smooth-AP) Using Resnet-50 (R-50) and Vitb-16 (V-16), Under  Original Loss (O-L) and Improved Loss (I-L), with Original Data (O-D) vs Augmented (I-D) Data Analysis, including Mean AP (mAP) of Each Kind. }
\begin{tabular*}{0.9\textwidth}{@{\extracolsep{\fill}} c c c c c c c c c c c}
\hline
& \multicolumn{2}{c}{Crrosion-AP} & \multicolumn{2}{c}{Dent-AP} & \multicolumn{2}{c}{Scratch-AP} & \multicolumn{2}{c}{Smooth-AP} & \multicolumn{2}{c}{mAP} \\
\cline{2-11}
& R-50 & V-16 & R-50 & V-16 & R-50 & V-16 & R-50 & V-16 & R-50 & V-16 \\
\hline
A-D/O-L & 70.05 & 71.84 & 71.72 & 79.75 & 65.53n & 63.45 & 81.60 & 82.61 & 72.23 & 74.41 \\
A-D/I-L & \textbf{81.61} & \textbf{80.48} & \textbf{87.21} & \textbf{87.38} & \textbf{82.53} & \textbf{79.49} & \textbf{95.91} & \textbf{96.81} & \textbf{86.82} & \textbf{86.04} \\
O-D/I-L & 33.46 & 33.69 & 35.29 & 34.67 & 33.29 & 30.63 & 33.67 & 38.79 & 31.11 & 37.36 \\
\hline
\label{tab : res}
\end{tabular*}
\end{table*}
\begin{figure*}[hbtp]
    \centering
    \includegraphics[width=0.8\linewidth]{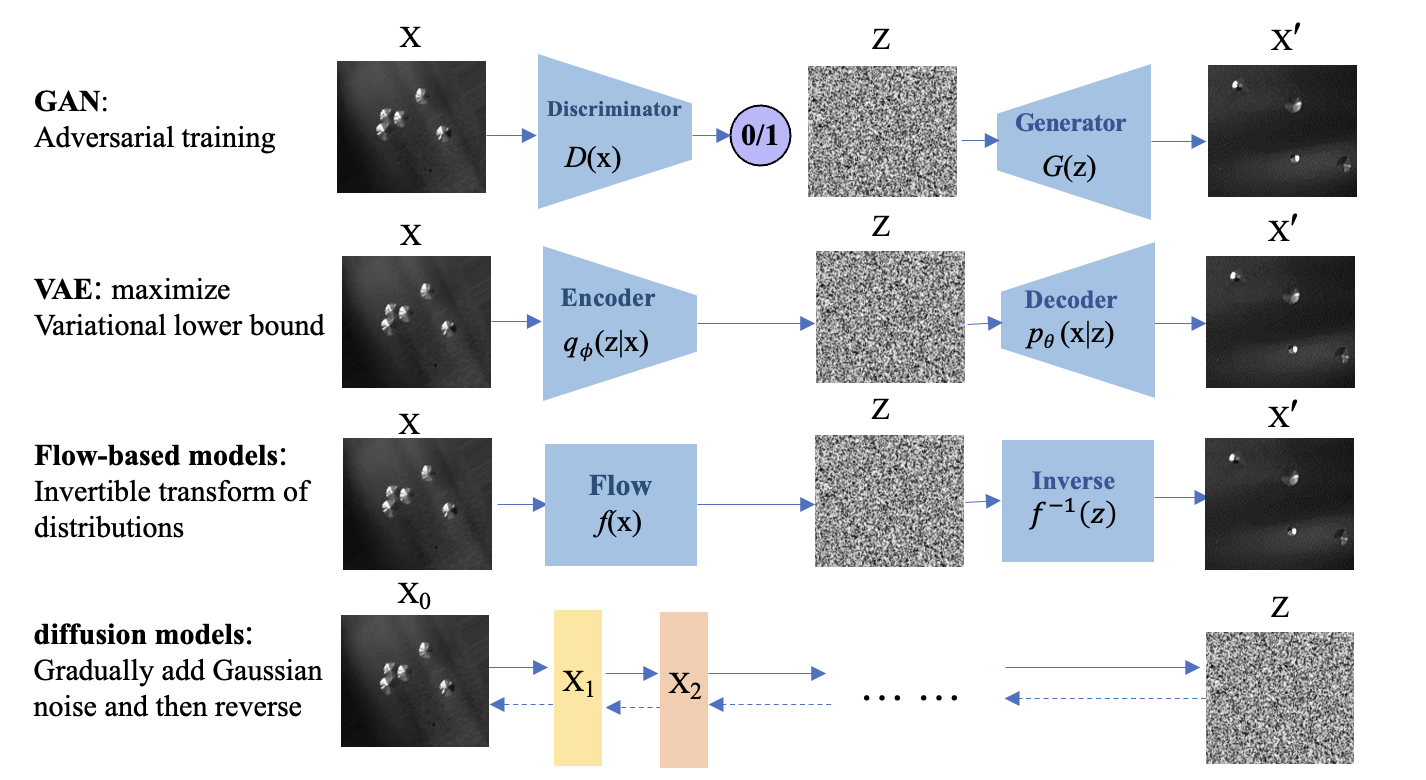}
    \caption{Overview of examples of aluminum defect images generated by common generative models.}
    \label{fig:fig15}
\end{figure*}
\noindent\textbf{\textit{3) Results on argumented data}}\\
We used two model frameworks, Resnet-50 and Vision Transformer B-16 (Vitb-16), to evaluate the Average Precision (AP) for four categories (erosion, dent, scratch, smooth) under original loss (I-L) and improved loss (D-L), as well as with original versus augmented data. The results in Table~\ref{tab : res} showes that transitioning from original to improved loss notably improves AP values; for example, AP in the erosion category increases from 72.12 to 79.97 with Resnet-50, and from 70.02 to 82.87 with Vitb-16. Further analysis reveals that augmented data provided higher AP values under improved loss conditions than original data, with Resnet-50 and Vitb-16 achieving 79.97 and 82.87, respectively, significantly outperforming the original data's 33.46 and 33.69. These results underscore the significant impact of data augmentation on enhancing model performance in complex prediction tasks and confirm that both models achieve optimal comprehensive performance under combined conditions of augmented data and improved loss.

\subsection{Visualization results} 
As shown in Figure~\ref{fig:fig14}, the first row displays defect image samples from the real world, characterized by their high realism. The second row features our augmented images, which are primarily synthetic and created using the diffusion model based on the original data. These synthetic images surpass the originals in terms of color fidelity, realism, texture, and perceptual quality, and are also more numerous.  Additionally, as detailed in Figure~\ref{fig:fig15}, we evaluated the outputs from four generative models: GAN, VAE, Flow-based model, and Diffusion model. Our analysis demonstrates that the Diffusion model excelled among them, producing defect images of superior quality with notable textures and enhanced details, thus delivering higher perceptual quality.

\textbf{\section{Conclusion}} 
This paper presents the DDPM-MoCo framework, addressing the issue of limited training data for defect detection in aluminum plates. It also enhances defect learning and overall contrastive performance by improving the contrastive loss for ResNet50 and ViT-B-16 backbone models in contrastive learning. Additionally, integrating neck and head modules from ResNet50 and ViT-B-16, with extra supervised training, boosts downstream prediction performance. In future work, we aim to demonstrate the framework's effectiveness in generating abundant high-quality images across various defect types and materials. We also plan to validate DDPM-MoCo's efficacy in enhancing feature extraction across a broader range of backbone models and also in improving the prediction accuracy by integrating corresponding neck and head modules into backbone models through further supervised training. \\ \\

%%%%%%%%% REFERENCES
{\small
\nocite{*}
\bibliographystyle{named.bst}
\bibliography{ijcai24}
}

\end{document}